\documentclass[letterpaper]{article} % DO NOT CHANGE THIS
\usepackage{aaai23}  % DO NOT CHANGE THIS
\usepackage{times}  % DO NOT CHANGE THIS
\usepackage{helvet}  % DO NOT CHANGE THIS
\usepackage{courier}  % DO NOT CHANGE THIS
\usepackage[hyphens]{url}  % DO NOT CHANGE THIS
\usepackage{graphicx} % DO NOT CHANGE THIS
\usepackage{natbib}  % DO NOT CHANGE THIS AND DO NOT ADD ANY OPTIONS TO IT
\usepackage{caption} % DO NOT CHANGE THIS AND DO NOT ADD ANY OPTIONS TO IT
\usepackage{booktabs}  
\usepackage{threeparttable} 
\usepackage{multirow}
\usepackage[utf8]{inputenc}
\usepackage{graphicx}
\usepackage{color}
\usepackage{xcolor}
\usepackage{amssymb}
\usepackage{amsmath}
\usepackage{makecell}
\usepackage{pythonhighlight}

\usepackage[colorlinks,
             linkcolor=black,
             anchorcolor=black,
             citecolor=black]{hyperref}
\usepackage{algorithm}
\usepackage[noend]{algorithmic}
\usepackage{newfloat}
\usepackage{listings}
\DeclareCaptionStyle{ruled}{labelfont=normalfont,labelsep=colon,strut=off} % DO NOT CHANGE THIS
\floatstyle{ruled}
\newfloat{listing}{tb}{lst}{}
\floatname{listing}{Listing}
\usepackage[switch]{lineno}
\usepackage{xspace}
\usepackage{colortbl}
\usepackage{stfloats}
\newcommand{\simmc}{\textsc{Simmc}\xspace}
\newcommand{\simmco}{\textsc{Simmc 1.0}\xspace}
\newcommand{\simmct}{\textsc{Simmc 2.0}\xspace}
\newcommand{\name}{\textsc{Spring}\xspace}

\def\eg{\emph{e.g.}}
\def\ie{\emph{i.e. }}

\newcommand{\comments}[1]{\STATE \small {\color{gray} /*\;\tt{\scriptsize #1}\;*/}}

\title{SPRING: Situated Conversation Agent Pretrained with \\ Multimodal Questions from Incremental Layout Graph}
\author {
    Yuxing Long \textsuperscript{\rm 1}, 
    Binyuan Hui \textsuperscript{\rm 2},
    Fulong Ye \textsuperscript{\rm 1}, 
    Yanyang Li \textsuperscript{\rm 2},
    Zhuoxin Han \textsuperscript{\rm 1}, \\ 
    Caixia Yuan \textsuperscript{\rm 1}, 
    Yongbin Li \textsuperscript{\rm 2}, 
    Xiaojie Wang \textsuperscript{\rm 1 \footnote{Corresponding author}}
}
\affiliations {
    \textsuperscript{\rm 1} Beijing University of Posts and Telecommunications, Beijing, China \\
    \textsuperscript{\rm 2} Independent Researcher \\
    \texttt{\{longyuxing, fulong\_ye, hanzhuoxin, yuancx, xjwang\}@bupt.edu.cn}\\ 
    \texttt{lyb821@gmail.com}
}

\usepackage{bibentry}
\begin{document}
% \linenumbers
\maketitle

\begin{abstract}
Existing multimodal conversation agents have shown impressive abilities to locate absolute positions or retrieve attributes in simple scenarios, but they fail to perform well when complex relative positions and information alignments are involved, which poses a bottleneck in response quality. In this paper, we propose a \textbf{S}ituated Conversation Agent \textbf{PR}etrained with Multimodal Questions from \textbf{IN}cremental Layout \textbf{G}raph (\textbf{\name}) with abilities of reasoning multi-hops spatial relations and connecting them with visual attributes in crowded situated scenarios. Specifically, we design two types of Multimodal Question Answering (MQA) tasks to pretrain the agent. All QA pairs utilized during pretraining are generated from novel Incremental Layout Graphs (ILG). QA pair difficulty labels automatically annotated by ILG are used to promote MQA-based Curriculum Learning. Experimental results verify the \name's effectiveness, showing that it significantly outperforms state-of-the-art approaches on both \simmco and \simmct datasets.
% (\href{https://github.com/LYX0501/SPRING}{https://github.com/LYX0501/SPRING})
We release our code and data at \href{https://github.com/LYX0501/SPRING}{https://github.com/LYX0501/SPRING}.
\end{abstract}

\section{Introduction}
Building multi-modal conversation agents that can communicate with people in visual situations is an attractive goal for the AI community. Lots of specific tasks and datasets for visual dialog, like VisDial \cite{visdial}, GuessWhat \cite{guesswhat}, GuessWhich \cite{guesswhich}, 
% and MMD \cite{mmd}, 
are proposed in recent years. Among them, the Situated Interactive Multi-modal Conversation (\simmco) \cite{simmc1} aims to study task-oriented dialogues that encompass a situated multi-modal user context in the form of a virtual reality (VR) environment. The updated dataset \simmct \cite{simmc2} provides a more challenging test bed for multi-modal conversation agents. There are many assets with a complex layout in each image. Figure 1 gives an example of a scene and a fragment of dialogue in \simmct. There are dozens of clothes in the image. Each cloth is a digit asset with a unique asset ID and a set of attributes (\eg{\,type, color}) in the metadata. But there is no information on the scene layout except for a few labels on four relations (up, down, left, and right) between the assets. 

\begin{figure}[htp]
    \centering
    \includegraphics[width=0.9\linewidth]{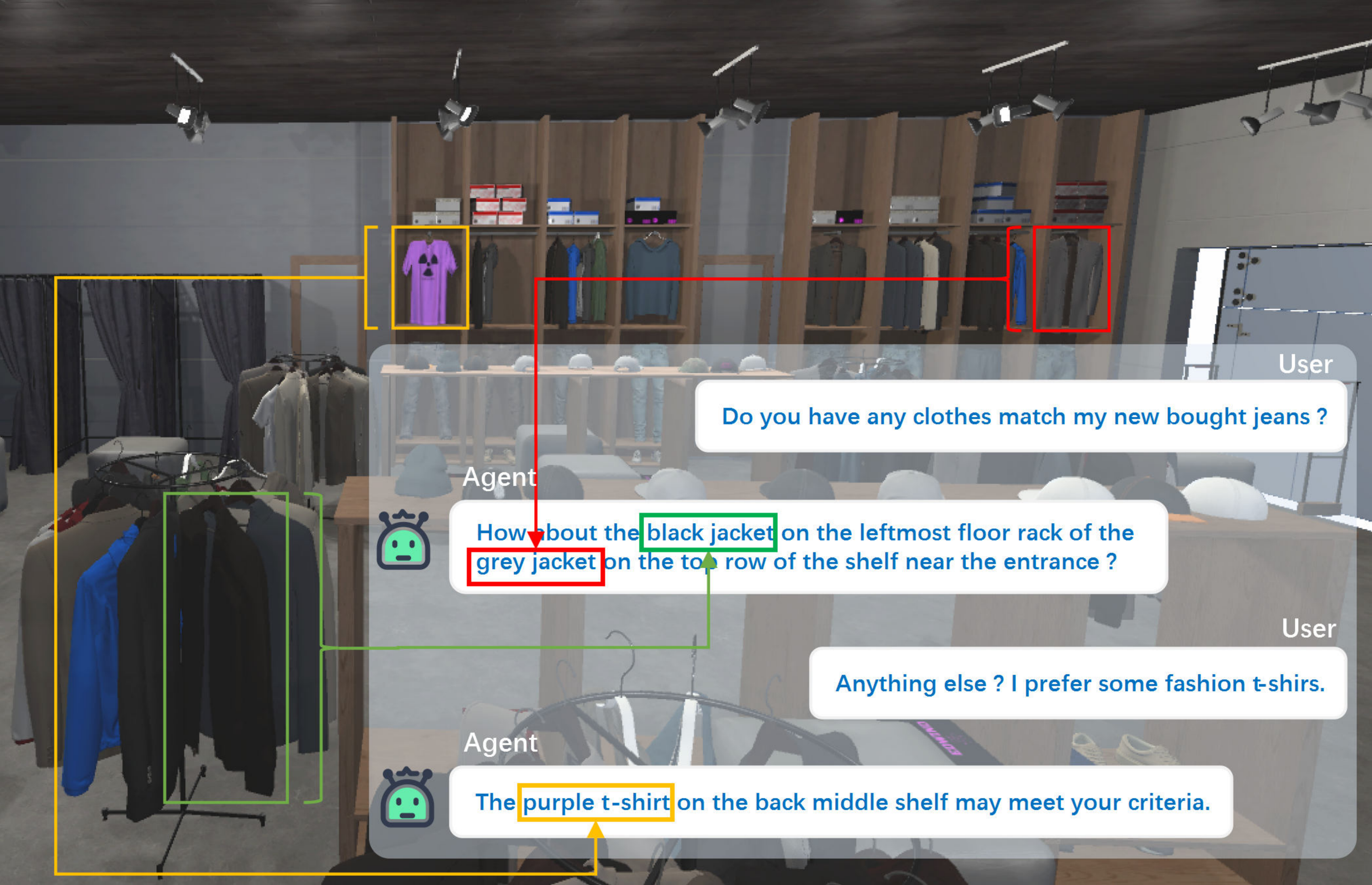}
    \caption{An example of a virtual scene and a fragment of dialogue  in \simmct. Since there are many clothes with similar visual attributes, it is difficult to talk about a asset only by its visual attributes. For this case, the spatial relations in the layout are necessary.}
    \vspace{-0.4cm}
    \label{fig:Intro}
\end{figure}

A number of works have been established on \simmct. Based on different multi-modal Visual-Language pre-training Models (VLM), previous researchers pay more attention to learning the visual attributes of assets. QS Goal Diggers \cite{simmc2overview} and Kakao Enterprise \cite{kakao} directly insert visual attributes into models input, while Sogang University \cite{simmc2} and A-STAR \cite{astar} build a set of visual attributes prediction tasks in pre-training stage. KAIST \cite{kaist} designs an auxiliary task to predict visual attributes. However, less attention has been paid to building spatial relations between assets. All existing models only use the coordinates of asset bounding boxes as positional information, which cannot capture spatial relations in the scenes.

As a result, the models can accurately generate \textit{"the black jacket"} but fail to describe more natural referring expressions like \textit{"the black jacket on the leftmost floor rack"}. It is obvious that the later expression is more useful in a scene including lots of clothes with similar attributes. The combination of attributes and spatial relations helps people locate assets quickly. To generate this type of expression, a model needs to learn not only the visual attributes of each asset but also spatial relations between different items.  

To address above problem, we propose \textbf{S}ituated Conversation Agent \textbf{PR}etrained with Multimodal Questions from \textbf{IN}cremental Layout \textbf{G}raph (\textbf{\name}), which is pretrained with multimodal questions generated from incremental layout graph. In our method, we design \textbf{Incremental Layout Graph (ILG)} for each scene to capture rich spatial relations between different scene items. Unlike scene graph \cite{sgsurvey}, an ILG is built using pure textual information and can be extended incrementally with newly added dialogue. And then, two types of \textbf{Multimodal Question Answering (MQA)} pre-training tasks and corresponding QA pairs are collected by traversing nodes (digital assets and background items) on the ILG. According to the spanned path length, QA pairs can be automatically annotated with difficulty levels. Curriculum Learning \cite{cl} is therefore employed for pre-training on a Transformer \cite{transformer} encoder-decoder backbone. Experiments on both \simmco and \simmct show that our method improves the response quality by a large margin compared to previous best models.

The main contributions of our work are as follows: 
\begin{itemize}
    \item[$\bullet$] We first propose a novel approach to build ILGs for virtual scenes from dialogue text incrementally. The ILGs include scene items with relations. It is worth noting that this process does not rely on any human annotation.
	\item[$\bullet$] Based on ILGs, we introduce two types of new MQA pretraining tasks that can facilitate model understanding of visual metadata and spatial relations between different assets. Pre-training samples are automatically generated by traversing the ILG, which also generates an accompanying difficulty label for curriculum learning.
	\item[$\bullet$] We conduct thorough experiments to verify that our approach effectively enhances response quality. Our approach outperforms existing state-of-the-art methods by a significant margin consistently on all metrics on both \simmco and \simmct.
\end{itemize}

\section{Related Works}

\paragraph{Situated Interactive Multimodal Conversations.}
Conversation systems have developed rapidly in recent years, \eg, task-oriented conversations pretraining \cite{he2022galaxy,he2022tree,he2022unified}, knowledge-based conversations \cite{s2sql,Wang2022ProtonPS} and so on. Among them, multimodal conversations are the new trend. 
META releases multimodal conversation datasets SIMMC \cite{simmc2} based on VR shopping stores. There are hundreds of scene snapshots from different angles. 
Compared with the previous multimodal dialogue datasets MMD \cite{mmd} and VisDial \cite{visdial}, the situated agent is required to generate more complex visual attributes and more detailed spatial relations to infer digital assets in the scene.
\citet{simmc2overview} has preliminary explorations on utilizing visual attributes and spatial relations. 
Concretely, DialVinVL \cite{simmc2overview} incorporates slot values about visual attributes with dialogue history as textual input and concatenates original box coordinates to region features extracted by the object detector as visual input. 
JMGPT \cite{simmc2} and JointGM \cite{astar} apply language model to predict visual attributes and system response jointly. 
MMBart \cite{kaist} adds embedded box coordinates to textual embedding as Transformer input and designs auxiliary tasks to predict visual attributes according to the output of encoder hidden states. We can find that their utilized spatial information is all from the bounding box.
Unlike these methods, we first notice the lack of VLM's capability for visuality and spatiality, and then propose MQA pretraining tasks based on incremental layout graphs which have been successfully applied to \cite{SocAoG,Liao2021DialogueST,Hui2021DynamicHR,Qiu2022TowardsSI}. 

\paragraph{Visual Language Pretraining.}
To improve models' perception of text and image and help them establish connections between multimodal information, kinds of visual language pretraining models are designed. ViLBERT \cite{vilbert} and UNITER \cite{uniter} propose to consider the raw output of the detector, a distribution of object classes, as soft labels and optimize the KL-divergence between two distributions. LEXMERT \cite{lxmert} and UNIMO \cite{unimo} propose Masked Region Feature Regression (MRFR) regresses the masked region feature to its corresponding original region feature, where represents images as a sequence of region features by Faster R-CNN \cite{faster-rcnn}. 
Furthermore, SOHO \cite{soho} is designed to avoid information leakage from neighbor features when images are converted into grid features or patch features.

Recently, CLIP \cite{clip} and ALIGN \cite{align} leverage large-scale image-text pairs to learn transferable visual representations and exhibit surprising zero-shot transfer to image classification tasks. VL-T5 \cite{vlt5} and OFA \cite{ofa} introduce downstream tasks, like visual grounding and grounded caption, into pretraining tasks to narrow the gap between pretraining and fine-tuning. 
Unlike these efforts, we design new pretraining tasks through a unified QA paradigm to improve existing methods' visual attributes and spatial relations modeling without adding new modules.

\begin{figure*}[htp]
    \centering
    \includegraphics[width=0.8\linewidth]{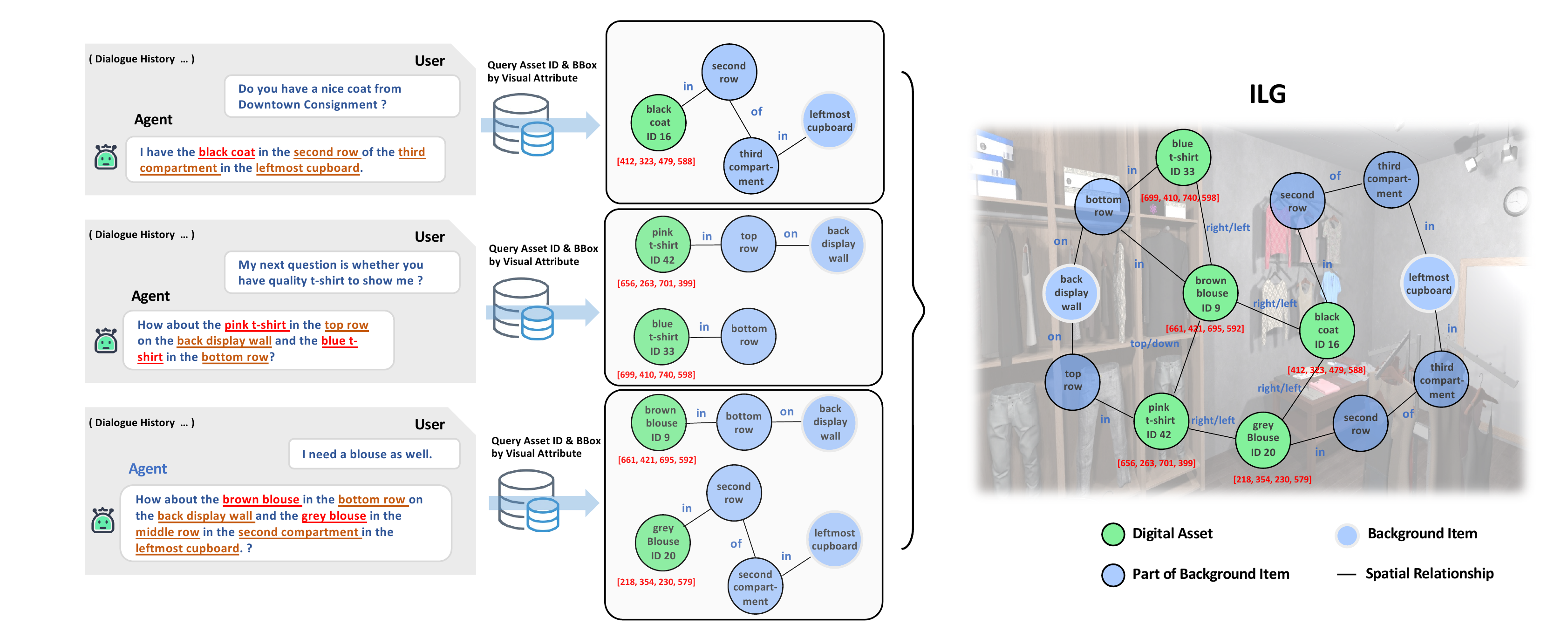}
    \caption{Construction of Incremental Layout Graph from dialogue. Digital assets and background items constitute ILG nodes while spatial relations form ILG edges. ILG is continuously incremented with newly added dialogue under the same scene.}
    \vspace{-0.5cm}
    \label{fig:ILG}
\end{figure*}

\section{Methods}
\iffalse
We denote a dialog with $T$ turns: $\mathcal{D}=\left\{\left(U_{i}, R_{i}, I_{i}\right)\right\}_{i=1}^{T}$, where $U_i$ and $R_i$ are the user and system utterances at turn $i$, $I_i$ is the scene image at turn $i$. Given such a dialog $\mathcal{D}$ sample, at the final turn $t$ ($t$ = $T$), with the current user utterance $U_{t}$, the dialog history $H_{t}=\left[U_{j}, R_{j}\right]_{j=1}^{t-1}$ and the scene image $I_{t}$, the task is to predict the natural language system response $R_{t}$. The task can be concluded as follow.
\fi
Let $\mathcal{D}=\left\{\left(U_{t}, R_{t}, I_{t}\right)\right\}_{t=1}^{T}$ be an ongoing dialogue between a user and an agent with $T$ rounds, where $U_t$ is the user utterance at time step $t$, $R_t$ is the language response to $U_t$ by the agent, and $I_t$ is the accompanying scene image. The task here is to predict the optimal language response $R_{t}^{'}$, given the dialog history $H_{t}=\left[U_{i}, R_{i}\right]_{i=1}^{t-1}$, current user utterance $U_{t}$ and scene image $I_{t}$, as modeled in Eq (1)
\begin{equation}
\resizebox{0.55\hsize}{!}{$
R_{t}^{'}=argmax \,  P_{\theta} \left(R_{t} \mid H_{t}, U_{t}, I_{t}\right)
$}
\label{eq1}
\end{equation}
where $\theta$ is the model learnable parameters.

To solve above problem, we propose a mutimodal dialogue model \textbf{\name}, which is pretrained with mutlimodal questions generated from incremental layout graph. In the following sections, we will introduce model architecture, ILG generation and MQA pretraining tasks in order.

\subsection{Architecture}
The backbone of \textbf{\name} model is encoder-decoder based single-stream VLM framework, which are stacks of Transformer \cite{transformer} layers. The scene image $I_t \in \mathbb{R}^{h \times w \times c}$ is splitted to $P$ patches. And each patch is projected to visual embedding of the model hidden size. The dialogue history and current user utterance are converted to sub-word sequence by Byte-Pair Encoding (BPE) and then embedded to textual embedding. All visual embedding and textual embedding are concatenated as model input.

To facilitate \textbf{\name} to better understand the information of embodied scenes, we propose a series of MQA  pre-training tasks based on layout graphs $\mathcal{G}_{i}$. As there is no annotated layout graph in the \simmc dataset, we propose an unsupervised ILG construction method based on natural language dialog history.

\subsection{Incremental Layout Graph (ILG)}
We observe that visual attributes and spatial descriptions exist in the dialogue history. Compared with dataset annotations, the information from dialogue is more detailed. For example, \simmct annotation only gives bounding boxes of digital assets and four types of relative position (up, down, right, left) between them, while dialogues include the absolute position of background items and their relative position with assets. A crucial discovery lies in the co-coreference between dialogue history and response, the same assets present in the responses to be generated.

Therefore, we propose an ILG generation algorithm to extract high quality information from dialogues and generate Incremental Layout Graph (ILG) $\mathcal{G}_i = \langle \mathcal{V}_i, \mathcal{E}_i \rangle $ to 
dispose them, where $\mathcal{V}_i$ denotes the node set containing the digital assets and background items from dialog history and $\mathcal{E}_i$ represents the edge set depicting spatial relations between scene items $\mathcal{V}_i$.

\paragraph{Textual Information Extraction and Alignment}
we consider adopting a rule-based textual information extraction method, \ie, regular expression, to extract visual attributes and spatial descriptions from dialogue history without human annotation.
The regular expressions $\mathrm{RegExp}_{va}$ and $\mathrm{RegExp}_{sd}$ for \textbf{visual attribute} and \textbf{spatial description} are as follows.
\vspace{-0.05cm}
\begin{gather}
    \mathrm{RegExp}_{va} = (\mathrm{art}.)\ (\mathrm{color})\ (\mathrm{asset}\;\mathrm{type}) \\
    \mathrm{RegExp}_{sd} = (\mathrm{positional}\;\mathrm{prep}.)\ (\mathrm{art}.)\ (.*?)\ (\mathrm{punc}.)
\end{gather}
where $\mathrm{art}$. is article, $\mathrm{prep}$. represents preposition and $\mathrm{punc}$. means punctuation. Please refer to  Appendix for details. With these two regular expressions, as left part of Figure \ref{fig:ILG} shows, we can extract visual attribute \textit{"black coat"} and spatial description \textit{"in the second row of the third compartment in the leftmost cupboard"} from dialogue history.

Although the visual attributes and spatial descriptions extracted by the above regular expressions are naturally aligned because of language features, they are not aligned with asset IDs, making asset box coordinates unusable. To solve this problem, we query the color and type of assets from the database by their IDs to compose visual attributes like \textit{"black coat"} and then try pairing them with the extracted visual attributes like \textit{"black coat"}. If these two visual attributes match, the corresponding asset ID "16" can be determined, from which we can get the paired asset IDs, visual attributes, and spatial descriptions. We further design the following two regular expressions to extract \textbf{background item} and \textbf{relative spatial relation} from extracted spatial descriptions.
\vspace{-0.05cm}
\begin{gather}
    \mathrm{RegExp}_{bi} = (\mathrm{background}\;\mathrm{item}) \\
    \mathrm{RegExp}_{sr} = (\mathrm{positional}\;\mathrm{prep}.)
\end{gather}
where $\mathrm{RegExp}_{bi}$ and $\mathrm{RegExp}_{sr}$ denote regular expressions for background item and spatial relation, $\mathrm{prep}$. represents preposition. With these two regular expressions, as middle part of Figure \ref{fig:ILG} shows, we can extract background items \textit{"second row"}, \textit{"third compartment"} and \textit{"leftmost cupboard"} and relative spatial relations \textit{"in"}, \textit{"of"} from spatial description obtained previously.

\begin{figure*}[htp]
    \centering
    \includegraphics[width=0.72\linewidth]{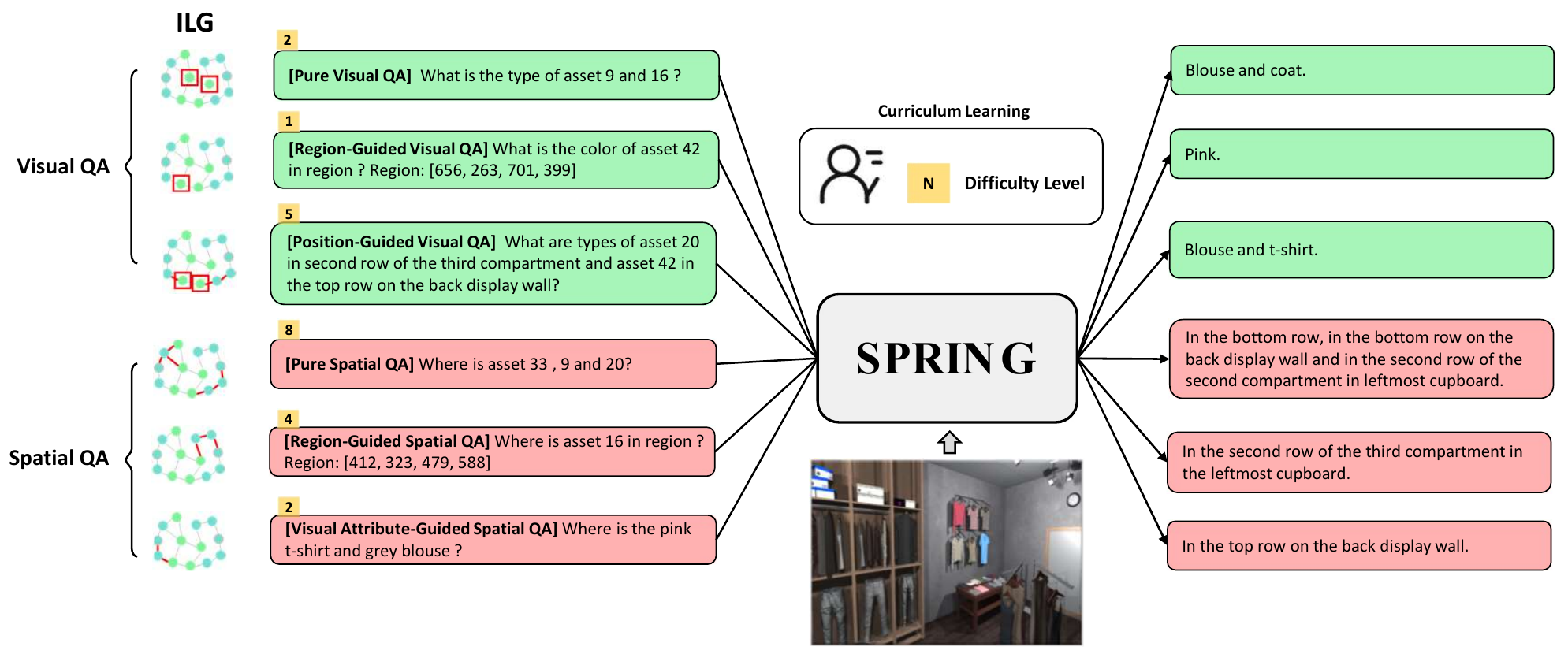}
    \caption{Demonstration of \name model and two types of MQA pretraining tasks, Visual QA and Spatial QA. Curriculum learning based on QA pair difficulty level  activates the potential of MQA tasks.}
    \label{fig:model}
\end{figure*}

\renewcommand\arraystretch{1.2}
\begin{table*}
    \centering
    \setlength\tabcolsep{4pt}
    \centering  
    \fontsize{6}{6}\selectfont  
    \begin{tabular}{llcccc}
        \toprule
        \textbf{\textsc{Scene Image}} & \textbf{\textsc{QA Type}} & \textbf{\textsc{Question Template}} & \textbf{\textsc{Answer}}  \\
        \midrule
         \multirow{4}{*}{\begin{minipage}[b]{0.3\columnwidth}
                		\centering
                		\raisebox{-.2\height}{\includegraphics[width=\linewidth]{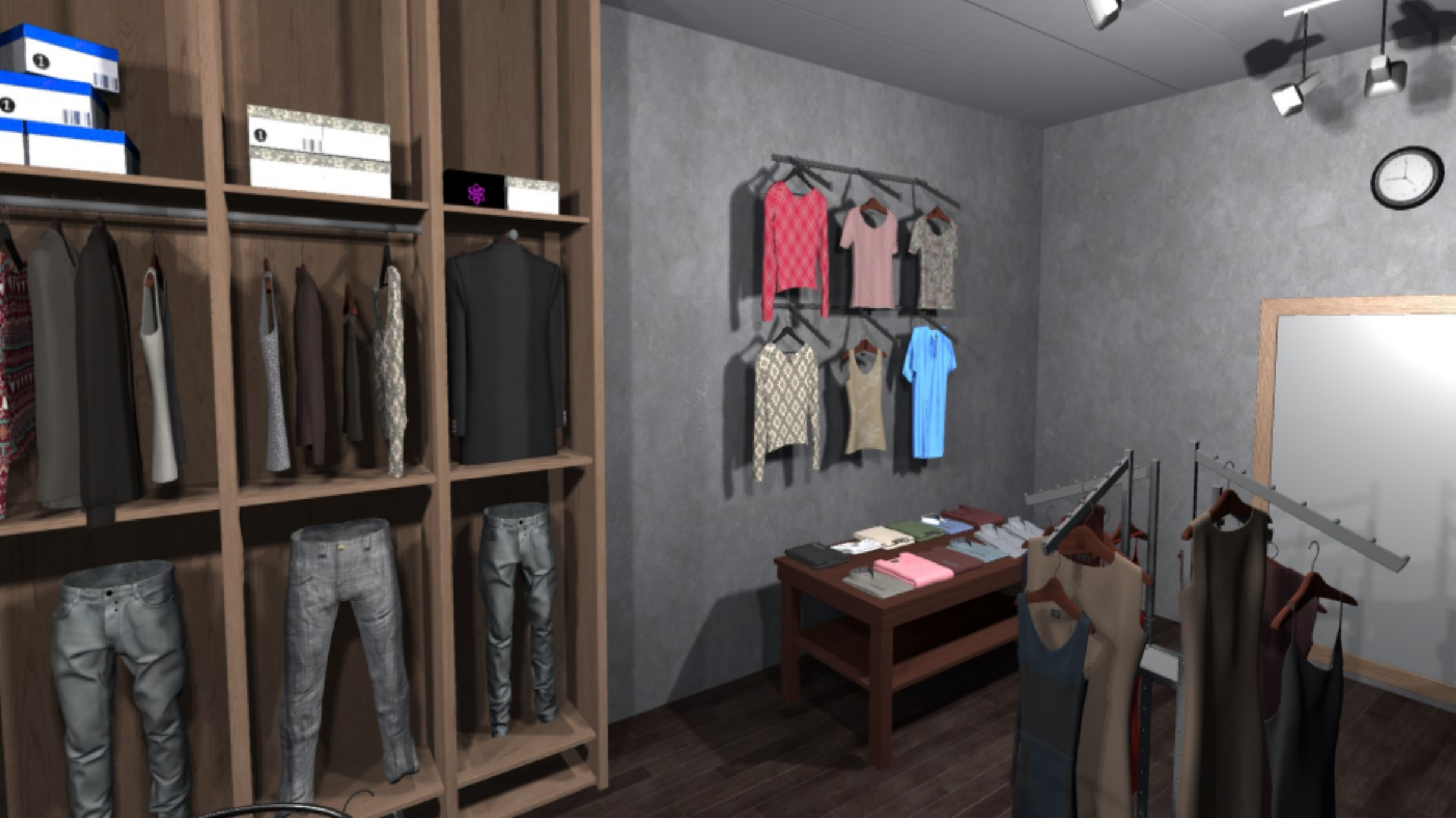}}
                	    \end{minipage}}   
         & \tt{Pure Visual QA}  & What is the \textbf{[visual attribute type]} of item \textbf{[asset ID]}?  & \textbf{[visual attribute value]}  \\ 
         & \tt{Region-Guided Visual QA}  & What is the \textbf{[visual attribute type]} of item \textbf{[asset ID]} in region? Region: \textbf{[x1, y1, x2, y2]}  & \textbf{[visual attribute value]} \\
         & \tt{Position-Guided Visual QA}  & What is the \textbf{[visual attribute type]} of item \textbf{[asset ID] [position]}?  & \textbf{[visual attribute value]} \\
         & \tt{Pure Spatial QA}  & Where is the item \textbf{[asset ID]}? & \textbf{[position]} \\
         & \tt{Region-Guided Spatial QA}  & Where is the item \textbf{[asset ID]} in region? Region: \textbf{[x1, y1, x2, y2]} & \textbf{[position]} \\
         & \tt{Visual Attribute-Guided Spatial QA} & Where is the \textbf{[item color] [item type] [asset ID]}?  & \textbf{[position]}\\
        \bottomrule
    \end{tabular}
    \caption{QA pair template. Square brackets ‘$\scriptsize [*]$’ represent slots to be filled by traversing ILGs.}
    \vspace{-0.5cm}
    \label{tab:qa_template}
\end{table*}

\paragraph{Incremental Layout Graph Generation}
With rich information extracted from a sample of dialogue history, layout sub-graph can be generated as middle part of Figure \ref{fig:ILG} shows. In the layout sub-graph, digital asset node store its visual attributes like \textit{"black coat"} and asset ID "16" while background item nodes store background items like \textit{"second row"}, \textit{"third compartment"} and \textit{"leftmost cupboard"}. Spatial relations and queried bounding boxes are utilized to define layout sub-graph edges. As the right part of Figure \ref{fig:ILG} shows, the scene ILG continuously increments with newly added sub-graph about the same scene, which finally can include all digital assets, background items, and spatial relations between them under this scene. Mining information on the ILG is simple but effective. The visual attributes can be easily obtained by traversing the ILG nodes, while multiple types of spatial relations can be inferred by walking along the ILG edges. 

\vspace{-0.05cm}
\subsection{ILG-based MQA Pre-training Tasks}
To enhance response generation quality of visual attributes and spatial relations, we design visual QA pre-training task and spatial QA pre-training task based on Multimodal Question Answering (MQA), which respectively contain three types of novel sub-tasks. As shown in Figure \ref{fig:model} and Table \ref{tab:qa_template}, all QA pairs are automatically generated by traversing ILG and filling the corresponding template. The QA pair generation algorithm is displayed in Algorithm \ref{algorithm:qa}. 
Formularly, we use the question template filling function $Q_{type}(\cdot)$ to generate question, 
$A_{type}$ represents corresponding answer, 
$\textrm{Type}_{va}$ means visual attribute type, $\textrm{ID}_{asset}$ denotes asset ID,  
$I_{scene}$ is scene image,
$\textrm{BBox}_{asset}$ means asset region coordinates,
$t_{sr}$ represents spatial relation,
$t_{va}$ is visual attribute, 
$t_{bi}$ denotes background item.

\subsubsection{\textbf{3.3.1} Visual QA}
\paragraph{Pure Visual QA (PVQA)}
As the most basic visual QA task, the goal of Pure Visual QA is to help the model establish connections between asset ID and corresponding visual attributes when a scene image is provided. We design PVQA template in which the question prompts the type of visual attribute and asset ID. 
The pure visual question can be generated by traversing the asset nodes of ILG and filling \texttt{\footnotesize [asset\,ID]} slot in the template while answers are generated based on the visual attributes stored in asset nodes. The objective of PVQA task is the following.
\begin{equation}
\resizebox{0.75\hsize}{!}{$
L_{\theta}=-\sum_{i=1}^{N} \log P_{\theta}\left(A_{pv} \mid Q_{pv}(\textrm{Type}_{va}, \textrm{ID}_{asset}), I_{scene}\right)
$}
\end{equation}

\paragraph{Region-Guided Visual QA (RVQA)}
To improve the model's ability of locating asset and describing its visual attribute by region visual context, we design RVQA template based on PVQA, in which the question is guided by asset region coordinates and asset ID. The region-guided visual question can be generated by traversing asset nodes of ILG and filling \texttt{\footnotesize [asset\,ID]}, bounding box coordinates \texttt{\footnotesize [x1,\,y1,\,x2,\,y2]} slots in the template. The corresponding answer is produced based on the visual attributes stored in asset nodes. The objective of RVQA task is the following.
\begin{equation}
\resizebox{0.85\hsize}{!}{$
L_{\theta}=-\sum_{i=1}^{N} \log P_{\theta}\left(A_{rgv} \mid Q_{rgv}(\textrm{Type}_{va}, \textrm{ID}_{asset}, \textrm{BBox}_{asset}), I_{scene}\right)
$}
\end{equation}

\begin{algorithm}[h]
\small
\caption{\textbf{QA Pair Generation}}
\begin{algorithmic}[1]
\REQUIRE ~~\\
ILG $\mathcal{G}_i = \langle \mathcal{V}_i, \mathcal{E}_i \rangle$, QA template list $T$
\ENSURE ~~\\
QA pair list $QA$, difficulty label list $DL$
\STATE Initialize QA pair list $QA$ and difficulty label list $DL$
\FOR{$node$ in $\mathcal{E}_i$}
\IF{$TypeOf(node) = "background\:item"$}
\STATE Skip $node$
\ENDIF
\comments{Get information from digtal asset node} \\
\STATE  $(t_{va}, ID_{asset}, BBox_{asset}) \leftarrow GetInfo(\mathcal{G}_i, node)$
\comments{Walk from node to get spatial relations} \\
\STATE $(t_{bi}, t_{sr}) \leftarrow Walk(\mathcal{G}_i, node)$
\STATE  $t_{slot} \leftarrow (t_{va}, t_{bi}, t_{sr}, BBox_{asset}, ID_{asset})$
\FOR{$template$ in $T$}
\comments{Fill in the template} \\
\STATE  $(qa, dl) \leftarrow FillIn(template, t_{slot})$
\STATE Add $QA \leftarrow qa, DL \leftarrow dl$
\ENDFOR
\ENDFOR
\end{algorithmic}
\label{algorithm:qa}
\end{algorithm}

\paragraph{Position-Guided Visual QA (PoVQA)}
In the conversations, instead of region coordinates, an agent has to locate asset by its spatial information no matter when understanding user utterances or making recommendations. To bring the  question closer to a real conversation, we design PoVQA template by replacing region coordinates in RVQA with spatial relations. For position-guided visual question template, the  \texttt{\footnotesize [asset\,ID]} slot can be filled by traversing asset nodes of ILG while the \texttt{\footnotesize[position]} slot is filled by spatial relation path between asset nodes and background item nodes. The corresponding answer is produced based on the visual attribute stored in asset nodes. The objective of PoVQA task is the following.
\begin{equation}
\resizebox{0.85\hsize}{!}{$
L_{\theta}=-\sum_{i=1}^{N} \log P_{\theta}\left(A_{pgv} \mid Q_{pgv}(\textrm{Type}_{va}, \textrm{ID}_{asset}, t_{sr}), I_{scene}\right)
$}
\end{equation}

% \vspace{-1cm}
\subsubsection{\textbf{3.3.2} Spatial QA}
\paragraph{Pure Spatial QA (PSQA)}
As the most basic spatial QA task, the goal of PSQA is to help the model establish connections between asset ID and corresponding spatial relations when a scene image is provided. We design PSQA template in which the question only prompts "where" and asset ID. The pure spatial question can be generated by traversing the asset nodes of ILG and filling \texttt{\footnotesize [asset\,ID]} slot in the template, while answers are generated based on the spatial relation paths between the background item node and the asset node. The objective of PSQA task is the following.
\begin{equation}
\resizebox{0.75\hsize}{!}{$
L_{\theta}=-\sum_{i=1}^{N} \log P_{\theta}\left(A_{ps} \mid Q_{ps}(\textrm{ID}_{asset}), I_{scene}\right)
$}
\end{equation}

\paragraph{Region-Guided Spatial QA (RSQA)}
To improve the model's ability of locating an asset and describing its spatial relations by region visual context, we design RSQA template based on PSQA, in which the question is guided by asset region coordinates and asset ID. The region-guided visual question can be generated by traversing asset nodes of ILG and filling the slots of \texttt{\footnotesize[asset\,ID]}, bounding box coordinates \texttt{\footnotesize [x1,\,y1,\,x2,\,y2]} in the template. The corresponding answer is produced based on the spatial relation paths between the background item node and the asset node. The objective of RSQA task is the following.
\begin{equation}
\resizebox{0.85\hsize}{!}{$
L_{\theta}=-\sum_{i=1}^{N} \log P_{\theta}\left(A_{rgs} \mid Q_{rgs}(\textrm{ID}_{asset}, \textrm{BBox}_{asset}), I_{scene}\right)
$}
\end{equation}

\paragraph{Visual Attribute-Guided Spatial QA (VSQA)}
In the conversations, instead of region coordinates, an agent has to locate an asset by its visual attribute no matter when understanding user utterances or making recommendations. To bring the question closer to a real conversation, we design VSQA template by replacing region coordinates in RSQA with visual attributes (\eg{\,color,\,type}). For spatial-guided visual question template, the \texttt{\footnotesize[asset\,ID]} slot can be filled by traversing asset nodes of ILG while the \texttt{\footnotesize [item\;color]} and \texttt{\footnotesize [item\;types]} slots are filled by the visual attribute stored in asset nodes. The corresponding answer is produced based on the spatial relation paths between the background item node and the asset node. The objective of VSQA task is the following.
\begin{equation}
\resizebox{0.85\hsize}{!}{$
L_{\theta}=-\sum_{i=1}^{N} \log P_{\theta}\left(A_{vags} \mid Q_{vags}(\textrm{ID}_{asset}, t_{va}), I_{scene}\right)
$}
\end{equation}

\begin{table*}
    \centering
    \scalebox{0.7}{
    \begin{tabular}{lccccccc|cc}
        \toprule
        %  {\textbf{\textsc{Models}}} & \multicolumn{7}{c|}{\textbf{\simmct}} & \textsc{\textbf{Visual}} & \textsc{\textbf{Spatial}} \\ &
         {\textbf{\textsc{Models}}} & \textbf{BLEU-1} & \textbf{BLEU-2} & \textbf{BLEU-3} & \textbf{BLEU-4} & \textbf{METEOR} & \textbf{ROUGE} & \textbf{CIDEr} & \textsc{\textbf{Visual}} & \textsc{\textbf{Spatial}} \\
        \midrule
        \multicolumn{10}{c}{\textbf{\simmco}} \\
        \midrule
        MN-MAG \cite{simmc1sogang1}    &27.28  &16.75   &12.32   &9.50   &16.62   &32.35   &0.8694    &9.49   &9.10  \\
        Tom \cite{simmc1sogang2}    &28.95  &18.81   &14.23   &11.10   &18.83   &38.18   &1.5014    &11.13   &10.17     \\
        JBi-encoder \cite{simmc1astar}   &26.76  &16.76   &12.49   &9.60   &17.65   &36.46   &1.2345    &9.73   &9.43     \\
         \rowcolor[RGB]{237,237,237} \name(Ours)    & \textbf{32.46}  & \textbf{22.15}   & \textbf{17.23}   & \textbf{13.77}  & \textbf{20.75}  & \textbf{40.51}  & \textbf{1.6329}  & \textbf{13.53}  & \textbf{12.60}   \\
        \midrule
        \multicolumn{10}{c}{\textbf{\simmct}} \\
        \midrule
         MTN  \cite{simmc2}  &62.38  &44.52   &32.90   &21.70   &21.38   &38.50   &1.1207    &19.91   &14.95     \\
         JMGPT  \cite{simmc2}  &51.05  &35.03   &24.66   &19.20   &14.73   &29.18   &0.7738    &13.67   &11.54    \\
         JMGPT-BS \cite{simmc2overview}   &64.86  &48.86   &37.91   &28.38  &22.43   &43.88   &1.9669    &22.10   &14.56      \\
         JointGM  \cite{astar}  &64.40  &48.54   &37.69   &34.62   &21.91   &42.44   &1.8265    &21.77   &15.82     \\
         MMBart \cite{kaist}   &69.89  &52.99   &41.32   &33.10   &24.79   &46.60   &2.1887    &26.19   &21.11     \\
         DialVinVL \cite{simmc2overview}    &75.38  &57.42   &44.92   &34.90   &27.09   &51.24   &2.3426    &29.92   &22.55     \\
         GPTDeIT \cite{kakao}   &68.43  &52.23   &40.95   &28.50   &24.81   &47.80   &2.2271    &25.04   &18.06     \\
         GLIMMeR \cite{glimmer}    &74.05  &56.85   &44.88   &35.31   &27.48   &50.92   &2.4952    &32.70   &22.58     \\
         \rowcolor[RGB]{237,237,237} \name(Ours)     & \textbf{83.29}  & \textbf{64.75}   & \textbf{52.41}   & \textbf{42.49}  & \textbf{31.90}  & \textbf{57.12}  & \textbf{3.1351}  & \textbf{38.87}  & \textbf{30.77}   \\
        \bottomrule
    \end{tabular}}
    \caption{Comparison on \simmco, \simmct dataset, visual and spatial subsets. Our model consistently outperforms strong baselines by a large margin on 7 widely-used metrics. Specially, evaluation on \textbf{Visual Subset} and \textbf{Spatial Subset} by BLEU-4 effectively verify the huge improvement of our model comes from better response about visual attribute and spatial relation.}
    \vspace{-0.5cm}
    \label{tab:simmc2}
\end{table*}

\subsection{MQA-Based Curriculum Learning}
\paragraph{Automatic Difficulty Level Annotation}
When generating QA pairs by walking on the ILG, the number of nodes spanned by the pathway can be recorded. The more nodes the path passes through, the more scene information contained in the corresponding QA pair, which means that the multimodal dialogue model needs more hops to make inferences. Therefore, we automatically label the difficulty level of each QA pair according to the number of nodes the path spans. For example, when generating the question \textit{“Where is the brown jacket 83 \& 1055?”} and the answer \textit{“it is on the floor rack near the entrance.”}, one asset node \textit{“brown jack 83 \& 1055”} and two background item nodes are spanned on the ILG. The difficulty level of this QA pair is annotated as 3. The following is the formal expression.
\begin{equation}
\resizebox{0.25\hsize}{!}{$
d = \frac{|V_{spanned}|}{D}
$}
\end{equation}
where $d$ denotes the normalized difficulty level of QA pair, $|V_{spanned}|$ represents the number of ILG nodes spanned by corresponding path, $D$ is the maximum value of ILG nodes spanned by the QA pair path in the dataset.

\paragraph{Pretraining Strategy}
With automatically annotated difficulty labels, we propose MQA based curriculum learning to activate the potential of our designed MQA pretraining tasks. We define the model competence $c$ as follows.
\begin{equation}
\resizebox{0.5\hsize}{!}{$
c(t)=\gamma \sqrt{\alpha \frac{t}{T}+\beta\left(1-\frac{t}{T}\right) \min^{2}(d)}
$}
\end{equation}
where $t$ is the index of current training step, $T$ represents the maximum number of training steps, $\min^{2}(d)$ means the minimum value of difficulty level $d$, $\alpha$ and $\beta$ are hyper-parameters, $\gamma$ is determined by $\alpha$ as $\sqrt{\frac{1}{\alpha}}$. Here we set $\alpha$ to 1.2 and $\beta$ to 0.8. At a given training step $t$, QA pair with difficulty smaller than or equal to $c(t)$ (\ie{$d \le c(t)$}) will be sampled for training. 
As such, our pretraining strategy focuses on QA pairs with lower difficulty in the early stage, aiming at helping the model form preliminary perception and inference capabilities for scene items. In the middle and late stages, more difficult QA pairs are added, which improves the model’s ability to generate visual attributes and spatial relations for multiple assets.

After MQA pretraining, \textbf{\name} model is fine-tuned on the \simmc response generation task. The auto-regressive language modeling objective is the following.
\begin{equation}
\resizebox{0.5\hsize}{!}{$
L_{\theta}=-\sum_{i=1}^{N} \log P_{\theta}\left(R_{i} \mid H_{i}, U_{i}, I_{i}\right)
$}
\end{equation}
where $N$ denotes the total number of training samples.

\section{Experiment}
\subsection{Set up}
\paragraph{Datasets.} To evaluate the performance of the proposed model, we first conduct experiments on widely-used situated multimodal dialogue datasets \simmco and \simmct. The \simmct dataset contains 7.2k fashion dialogs and 4k furniture dialogs, respectively. There are around 290 digital assets for fashion and 110 assets for furniture, which are rearranged within seed scenes to generate 160 different scenes. The \simmco dataset includes 6.6k fashion dialogs and 6.4k furniture dialogs. We evaluate model performance on the dev-test split of \simmco and \simmct, which has the same scale as the test-std \footnote{Not publicly available as a test set for the DSTC competition.} dataset.

In addition, we invite human experts to filter responses with visual attribute or spatial relation from \simmco and \simmct dev-test split to construct \textbf{Visual Subset} and \textbf{Spatial Subset}. We further evaluate models on these two subsets to prove the effectiveness of our model.
\vspace{-0.1cm}
\paragraph{Evaluation Metrics.} The official metric adopted by \simmct response generation task is BLEU-4, which only focuses on n-grams overlap between the predicted and target response. For a more comprehensive comparison, we add widely-used machine generation metrics: BLUE-n \cite{bleu}, METEOR \cite{meteor}, ROUGE \cite{rouge} and CIDEr \cite{cider} metrics. 
Compared with the accuracy based BLEU metric, METOR and ROUGH pay attention to recall and calculate how many n-grams from the target response exist in the predicted response, while CIDEr uses TF-IDF to assign larger weights to infrequent phrases.
\vspace{-0.1cm}
\paragraph{Implementation Details.}
Our model is based on Transformer \cite{transformer} structure with 12 layers, where ever Transformer block has 768 hidden units and 12 attention heads. Each patch is projected to features of the same size as the hidden units. We initialize \name parameters from pretrained VLM, \ie, OFA \cite{ofa}. During pretraining, our model is trained for 4 epochs with 8 batch sizes on 8 TESLA V100 GPU. Adam \cite{adam} is adopted as optimizer with a 4e-4 learning rate. Besides, the dropout rate is set to 0.2 to prevent over-fitting. During fine-tuning stage, we train 60 epochs on the \simmc train set with a learning rate of 4e-5 and a batch size of 16.

%\vspace{-0.2cm}
\paragraph{Compared Methods.} 
We compare \name with strong baseline methods from \simmco and \simmct.
On \simmco, MN-MAG \cite{simmc1sogang1} adopts a memory network as encoder and designs multimodal fusion gate to fuse information. 
Tom \cite{simmc1sogang2} esambles prediction results from several GPT-2 models. 
JBi-encoder \cite{simmc1astar} is jointly trained to predict belief state and response.
On \simmct, MTN \cite{mtn} separately encodes multimodal input while the visual encoder is guided by a query-aware attention encoder.
JMGPT \cite{simmc2} trains a multi-task GPT2-large, which takes dialogue history and flattened multimodal contexts as input.
Furthermore,  JMGPT-BS \cite{simmc2overview} extends JMGPT by inferring with different beam search sizes.
MMBart \cite{kaist} adds box coordinates embedding to textual input and proposes auxiliary tasks to predict asset attributes.
DialVinVL \cite{simmc2overview} is based on VinVL-Base \cite{vinvl}, concatenates original box coordinates to region features as visual input, and incorporates dialogue history with dialogue policy as textual input. 
GPTDeIT \cite{kakao} utilizes GPT2-large \cite{gpt2} as the text model to encode dialogue history and flattened slot values and DeIT-I \cite{deit} as the image model to encode assets referenced by current turn utterance. 
JointGM \cite{astar} leverages BART-large \cite{bart} to predict disambiguation label, belief state and response jointly according to inputted dialogue history.
Similar to GPTDeIT, GLIMMeR \cite{glimmer} also leverages GPT2-large and utilizes asset scene ID to help the model understand the semantics of each asset. Notably, GLIMMeR is the state-of-the-art method on \simmct and achieves the winner of the DSTC10.

\begin{table}
    \centering
    \scalebox{0.6}{
    \begin{tabular}{llccc}
        \toprule
        \textbf{\textsc{Task}} & \textbf{\textsc{Models}} & \textbf{\simmct} & \textsc{\textbf{Visual}} & \textsc{\textbf{Spatial}}  \\
        \midrule
        &VLM    &38.22   &34.67   &25.04  \\  
        \midrule
         \multirow{5}{*}{\textbf{Visual QA}}   
         & VLM  &  &   &      \\
         & \quad w/ PVQA   &40.75  &\cellcolor[RGB]{193,255,193} 36.54   &27.58    \\
         & \quad w/ RVQA    &41.27  & \cellcolor[RGB]{84,255,159} 37.02   &27.22      \\
         & \quad w/ PoVQA    &40.89  & 35.94   &28.08     \\
         & \quad w/ (PVQA + RVQA + PoVQA)   &41.36  & \cellcolor[RGB]{84,255,159} 37.59  &28.24      \\
         \midrule
         \multirow{5}{*}{\textbf{Spatial QA}}   
         & VLM  &  &   &      \\
         & \quad w/ PSQA    &41.18  &36.05   & \cellcolor[RGB]{255,218,185} 28.30     \\
         & \quad w/ RSQA    &40.77  &35.42   & \cellcolor[RGB]{255,218,185} 28.18     \\
         & \quad w/ VSQA    &40.40  &36.34   &27.97     \\
         & \quad w/ (PSQA + RSQA + VSQA)   &41.56  &36.25   & \cellcolor[RGB]{255,160,122} 28.49      \\
         \midrule
         \multirow{3}{*}{\textbf{All}}   
         & VLM  &  &   &     \\
         & \quad w/ all QA    &41.92  &38.52   &30.18  \\
         & \quad w/ (all QA + CL)    & \textbf{42.49}  & \textbf{38.87}   & \textbf{30.77}      \\
        \bottomrule
    \end{tabular}}
    \caption{Ablation study on \simmct dataset with BLEU-4 metric. Red and green shades represent a stronger advantage in the visual and spatial subsets, respectively.}
    \vspace{-0.5cm}
    \label{tab:ablation}
\end{table}

\subsection{Overall Performance}
Table \ref{tab:simmc2} displays the results of the model on the \simmco and \simmct dataset response generation task.
It can be seen that \name has exceeded previous models by a large margin and achieved state-of-the-art results on all representative machine generation metrics. 
On \simmct, \name is respectively 7.91, 7.33, 7.49, and 7.18 higher than previous best models on BLEU-n, varying n from 1 to 4. The significant increased percentage on BLEU-n manifests our model successfully utilizing more accurate words and phrases to make responses. Our model also shows excellent performance on recall-based metrics METEOR and ROUGE, of which the score improvements reach 4.42 and 6.2. When the CIDEr metric pays more attention to infrequent n-grams, \name still outperforms GLIMMeR with 0.64 on CIDEr. 
Besides, according to the right part of Table \ref{tab:simmc2}, our model exhibits the highest BLEU-4 scores on the visual subset and spatial subset, which verifies the improvement of our model is produced by its better understanding of visual attribute and spatial relation and ability to conduct reasoning with aligned information to generate more accurate responses.

\subsection{Detailed Analysis}

\paragraph{Ablation Study.}
As shown in Table \ref{tab:ablation}, we perform ablation experiments to evaluate the effectiveness of each pretraining task and curriculum learning strategy in \name. 
It can be observed that each MQA pretraining task brings significantly BLEU-4 improvement on the complete \simmct dataset compared with the basic VLM model. Specifically, VLM models pretrained with all visual QA tasks perform 2.92 higher than baseline on the Visual Subset, while VLM models pretrained with all spatial QA tasks display 3.45 improvement compared with baseline on the Spatial Subset, which can verify that visual QA and spatial QA respectively prompt model's ability of describing visual attribute and spatial relation. Besides, the last two rows further prove that our designed curriculum learning pretraining strategy effectively activates the potential of QA pretraining tasks and boosts model performance.

\paragraph{Human Evaluation.}

\begin{figure}
    \centering
    \includegraphics[width=0.75\linewidth]{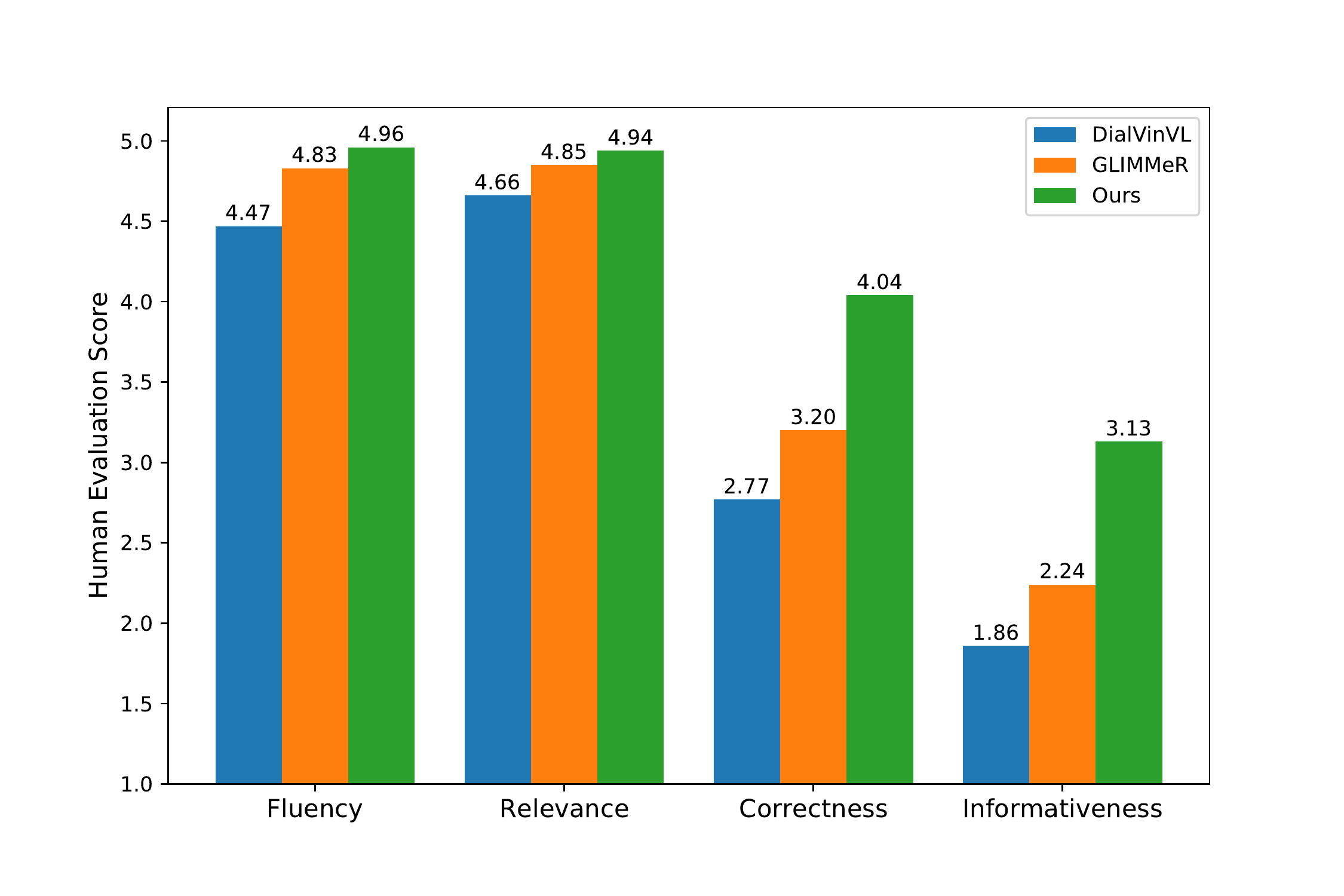}
    \caption{The human evaluation results on \simmct with four aspects. Our model displays significant improvement on \textbf{correctness} and \textbf{informativeness}.}
    \vspace{-0.15cm}
    \label{fig:human}
\end{figure}

The human evaluation mainly focuses on 4 aspects: \textbf{fluency}, \textbf{relevance}, \textbf{correctness}, and \textbf{informativeness}, which are important for task-oriented dialogue systems. We randomly select 500 dialogues from \simmct dev-test dataset as candidates, and then filter these dialogues from the results generated by DialVinVL, GLIMMeR, and our model. We release  evaluation task on Amazon Mechanical Turk (AMT) and make the last response of every selected dialogue evaluated by 10 different evaluators. Each evaluator scores 1500 generated responses on 4 aspects according to golden response in blind review from 1 to 5, simulating a real-life multimodal dialogue scenario. As shown in \ref{fig:human}, it can be observed that our model consistently outperforms the other two models on all metrics, which is in line with automatic evaluation results.

\begin{figure}[htp]
    \centering
    \includegraphics[width=1.0\linewidth]{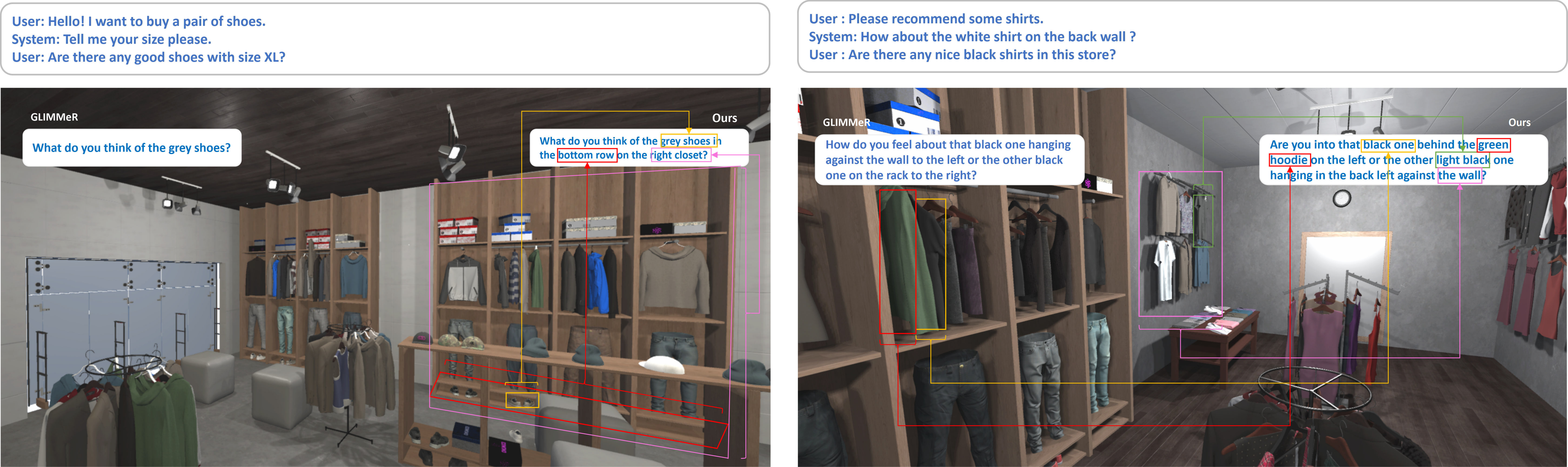}
    \caption{Case study on SIMMC 2.0 dataset.}
    \label{fig:case_study}
\end{figure}

\paragraph{Case Study.}
To better illustrate the advantage of our model and display how \name prompts model's ability of predicting visual attributes and spatial relations related to background items, we visualize several generated responses from our model and existing SOTA model with corresponding user utterance and scene snapshot, as shown in Figure \ref{fig:case_study}. It can be explicitly observed that: (1) our model is able to adopt background items to describe the position of target assets. (2) The relative spatial relations between target assets and background items can be accurately predicted by our model. (3) our model is equipped with the ability of aligning visual attribute to spatial information when multiple assets exist in the response.

\section{Conclusion}
In this paper, we propose a novel situated conversation agent pretraining method named \name. Specifically, all QA pairs and their difficulty labels used in pretraining are generated from our Incremental Layout Graph without any extra human annotations. 
Experimental results on \simmco and \simmct show that \name greatly surpasses previous models and describes visual attributes and spatial relations more accurately. 

\section*{Acknowledgement}
We would like to sincerely thank anonymous reviewers for their suggestions and comments. The work was partially supported by the National Natural Science Foundation of China (NSFC62076032). We also want to express our gratitude for precious advises given by Guanqi Zhan.

\bibliography{aaai23}

\clearpage
\section*{Appendix}

In the Table \ref{tab:RegExp_slot}, we display the slot types and slot values of four regular expressions we use to extract visual attribute, spatial description, background item, and spatial relation from dialogue corpus. All color values and type values are from \simmct metadata while background items are common furniture words provided by Wikipedia except for those appear in \simmct furniture types. The article, pronoun, coordinating conjunction and punctuation come from The Oxford English Dictionary. To make it 
clearer, we adopt a simple python code snippet to show how to use our designed regular expressions to extract visual and spatial information in the following.

\renewcommand\arraystretch{1.2}
\begin{table*}
    \centering
    \setlength\tabcolsep{4pt}
    \centering  
    \fontsize{6}{6}\selectfont  
    \begin{tabular}{llcccc}
        \toprule
        \textbf{\textsc{Slot Type}} & \textbf{\textsc{Slot Value}} \\
        \midrule
         \tt{fashion color}  & \makecell[l]{'red, white, yellow', 'red, white', 'purple', 'white, black', 'black', 'dark grey', 'grey, black', 'brown', 'dark green', 'grey', 'dark blue', 'blue, white',\\ 'grey, brown', 'white, blue', 'yellow, white', 'dark green, dark blue', 'light grey', 'white', 'blue', 'green', 'maroon', 'yellow',  'red', 'violet', \\'black, red, white', 'yellow, black', 'blue, black', 'black, white', 'light blue', 'red, black', 'pink, white', 'orange', 'yellow, brown', 'light pink', \\'dark brown', 'pink', 'dark yellow', 'light red', 'green, white', 'grey, white', 'black, red', 'grey, blue', 'brown, white', 'white, black, red', 'beige', \\ 'light orange', 'orange, purple', 'dirty green', 'blue, grey', 'black, grey', 'white, grey', 'olive', 'dark red', 'olive, black', 'pink, black', 'blue, green', \\ 'green, black', 'light blue, light green', 'dark pink, white', 'dirty grey', 'dark pink', 'red, grey', 'dark violet', 'olive, white', 'black, orange', 'golden', \\ 'maroon, white, blue', 'green, violet, pink', 'white, red, violet', 'brown, black', 'black, olive'}  \\
         \midrule
         \tt{fashion type}  & 'blouse', 'jacket', 'shirt', 'sweater', 'dress', 'tshirt', 'joggers', 'jeans', 'hat', 'tank top', 'vest', 'coat', 'shoes', 'skirt', 'suit', 'trousers', 'hoodie'  \\
         \midrule
         \tt{furniture color}  & 'red', 'blue', 'white', 'grey', 'brown', 'green', 'black', 'black and white', 'wooden'  \\
         \midrule
         \tt{furniture type}  & 'area rug', 'bed', 'chair', 'coffee table', 'couch chair', 'end table', 'lamp', 'shelves', 'sofa', 'table' \\
         \midrule
         \tt{background item}  & \makecell[l]{'rack', 'wall', 'mirror', 'shelf', 'closet', 'table', 'wardrobe', 'cabinet', 'window', 'divider', 'door', 'counter', 'cubby', 'cubbies', 'hanger', 'stand', \\ 'cupboard', 'mannequin', 'shoe boxes', 'room divider', 'wall divider'} \\
         \midrule
         \tt{positional preposition} & \makecell[l]{'in', 'on', 'at', 'behind', 'toward', 'to', 'against', 'of', 'along', 'below', 'towards', 'above'} \\
         \midrule
         \tt{article and pronoun} & \makecell[l]{'the', 'a', 'that', 'this', 'other', 'another'} \\
         \midrule
         \tt{punctuation and conjunctions} & ',', '.', ';', '?', 'and', 'or' \\
        \bottomrule
    \end{tabular}
    \caption{Slot types and slot values in the regular expression.}
    \vspace{-0.5cm}
    \label{tab:RegExp_slot}
\end{table*}

% \\ \hspace*{\fill} \\
\vspace{1cm}

\begin{python}
immport re

system_response = 'How about the blue
tshirt on the shelf or the red jacket 
above the table ?'

RegExp_vi = '(a|the|that|this|other|
another) (red, white, yellow|pink|
red, white|purple|white, black|black|
dark grey|light grey|white|blue|green|
maroon|yellow|red|violet|yellow, black|
black, red, white|blue, black|
black, white|light blue|red, black|
pink, white|orange|yellow, brown|
light pink|dark brown|pink|dark yellow|
light red|green, white|grey, white|
black, red|grey, blue|brown, white|
white, black, red|light orange|
orange, purple|dirty green|blue, grey|
black, grey|white, grey|olive|dark red|
olive, black|pink, black|blue, green|
green, black|light blue, light green) 
(blouse|jacket|shirt|sweater|dress|
tshirt|joggers|jeans|hat|vest|bed|
coat|shoes|skirt|suit|trousers|hoodie)'

RegExp_sd = '(in|on|at|behind|toward|
to|against|of|along|below|above) (the|
a|that|this|other) (.*?) (and|or|,|
\.|\?)'

RegExp_bi = '(rack|wall|mirror|shelf|
closet|table|wardrobe|cabinet|window|
divider|door|counter|cubby|cubbies)'

RegExp_sr = '(in|on|at|behind|toward|
to|against|of|along|below|towards|
above)'

extracted_vi = re.findall(RegExp_vi,
system_response)
# [('the', 'blue', 'tshirt'),
# ('the', 'red', 'jacket')]

extracted_sd = re.findall(RegExp_sd,
system_response)
# [('on', 'the', 'shelf', 'or'), 
# ('above', 'the', 'table', '?')]

sds = [' '.join(item[:-1]) for item
in extracted_sd]
# ['on the shelf', 'above the table']

bi_list, sr_list = [], []
for sd in sds:
    bi = re.findall(RegExp_bi, sd)
    bi_list.append(bi)
    sr = re.findall(RegExp_sr, sd)
    sr_list.append(sr)
# bi_list [('shelf'), ('table')]
# sr_list [('on'), ('above')]

\end{python}

\end{document}